# Bootstrapping A Wide-Coverage CCG from FB-LTAG

Christine Doran and B. Srinivas
Institute for Research in Cognitive Science, University of Pennsylvania
Philadelphia PA 19104-6228
email: {cdoran,srini}@linc.cis.upenn.edu

## Abstract

A number of researchers have noted the similarities between LTAGs and CCGs. Observing this resemblance, we felt that we could make use of the wide-coverage grammar developed in the XTAG project to build a wide-coverage CCG. To our knowledge there have been no attempts to construct a large-scale CCG parser with the lexicon to support it. In this paper, we describe such a system, built by adapting various XTAG components to CCG. We find that, despite the similarities between the formalisms, certain parts of the grammatical workload are distributed differently. In addition, the flexibility of CCG derivations allows the translated grammar to handle a number of "non-constituent" constructions which the XTAG grammar cannot.

## 1 Introduction

Our goal in undertaking this project is to develop a wide-coverage CCG system. Currently, a number of small-scale CCGs and parsers exist, but, to our knowledge, there have been no attempts to construct a large-scale CCG parser with the lexicon to support it. On the other hand, the XTAG project at University of Pennsylvania [2] currently has an implemented wide-coverage lexicalized LTAG system. In this paper we discuss the bootstrapping of a wide-coverage CCG from the XTAG grammar.

## 2 CCG and LTAG

A number of researchers have noted the similarities between LTAGs and CCGs. They have equivalent weak generative capacity, and have both been shown to belong to the class of mildly context sensitive languages, which is regarded as an interesting class for investigating natural language phenomena [6].[1] In addition to their weak equivalence, the elementary trees of LTAG are very similar to CCG derivations, if one views the trees as "structured types" [4]. For example, the simple transitive LTAG tree can be viewed as a function NP x NP → S, as can the CCG category (S\NP)/NP. One major difference between the two is that the LTAG trees represent a rigid structure, while CCG categories allow more flexibility in the derivation process. This flexibility gives CCG its well-known advantages over other formalisms in handling "non-constituent" constructions.

Observing these similarities between LTAG and CCG, we felt that we could build a CCG grammar from the XTAG grammar more efficiently than we could build it either from scratch or from a grammar based on some other formalism. We chose the following CCG combinators in order to match the grammatical coverage of the XTAG grammar: forward and backward application, forward and backward composition, backward crossing composition[2], and type-raising. The CCG system will use many modules from the XTAG system with few or no modifications. Figure 1 shows, in solid lines, the components of the XTAG system which are being translated into CCG; the components in dotted lines are being used unchanged.[3]

## 3 The Conversion Process

### 3.1 Trees to Categories

The first step in building the CCG system was to map LTAG trees to CCG categories. We chose 200 trees used in an XTAG-parsed corpus of 6,000 Wall Street Journal sentences as the initial set of trees to be translated (out of a total of about 566

---

[1] Lexicalization and features as used in this version of TAG have been shown not to increase the formal power of TAG.

[2] We need this combinator to handle particle shift, which requires a tree for each order in the XTAG grammar. This combinator also allows the CCG to handle heavy-NP shift, such as *Paddington loves dearly his very sticky marmalade sandwiches.*

[3] See Doran, et al. in this volume for information on the specifics of the XTAG system.

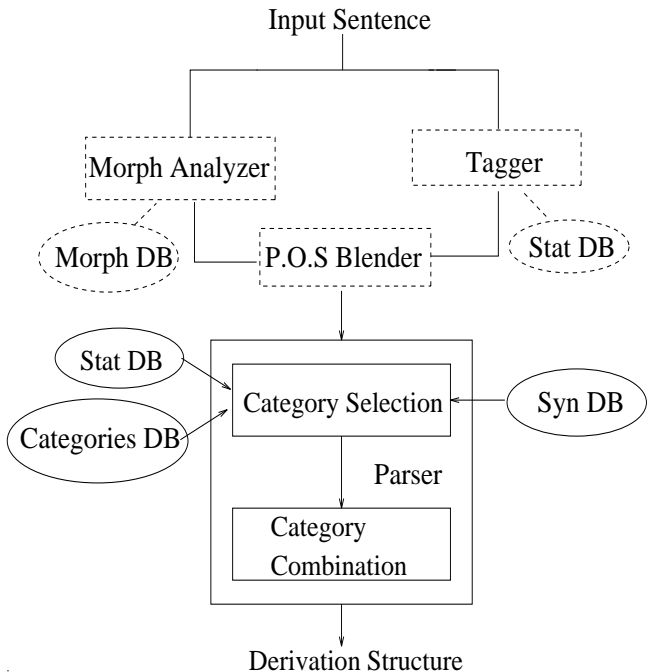

Figure 1: Components of the XTAG system modified to parse CCG

XTAG trees). As it turned out, the categories generated in this stage of the translation plus a few later additions actually covered just over 400 of the XTAG trees. These categories were used to build the CCG syntactic lexicon, which is compiled into the syntactic database (Syn DB) shown in the Figure 1. It contains a total of 124,517 pairings of lexical items and categories. The 78 translated categories are stored in the category database, which is indexed by the part-of-speech of the selecting word.

The basic approach to mapping an individual tree (say, a ditransitive tree) to a category is to: (1) take the root label of the XTAG tree as the result of the CCG category, $S$; (2) take the subject as the first argument, appearing to the left of the anchor $(S\backslash NP)$; (3) add each internal argument of the anchor as an argument of the category being constructed, starting with the outermost, $((S\backslash NP)/PP)/NP)$. Note that the anchor of the tree is not explicitly present in the category, and that the category contains sub-categories which may or may not be present as constituents in the XTAG tree. These differences in internal structure have repercussions in the CCG grammar.

### 3.2 Feature Mapping

Of the full set of 38 features used in the XTAG system, we selected a subset of the 13 which we considered the most central, for use in the CCG system.[4] Features come from three sources in the XTAG grammar. The first is from the morphology, where features such as number and case are associated with lexical items. This component is used as is from the XTAG system. The second source of features is the XTAG syntactic lexicon, where lexically specific features are instantiated. We translated only those features required by the CCG lexicon. The third is from the trees themselves, which use features to specify grammatical information holding generally of the tree regardless of which lexical item selects it. These features were mapped along with the trees, and are associated with categories in the category database. A sample entry from the CCG syntactic database is shown in Figure 2, with the corresponding category database entry.

(a) INDEX: park/2
    POS: V
    CAT: $(S0\backslash NP0)/NP1$   $(NP0\backslash NP1)/NP2$

(b) V: $(S0\backslash NP0)/NP1$  #INTRANS  #NP1caseacc
       $(NP0\backslash NP1)/NP2$  #INTRANSger  #NP2caseacc

Figure 2: (a): Syntactic DB, (b): Category DB

## 4 The Grammar

The most interesting aspect of translating LTAG to CCG is that, while they have the same formal power and superficially similar representations, certain parts of the grammatical workload are distributed differently. Some sets of XTAG trees collapse into single categories, while others multiply into large numbers of categories. For instance, each XTAG tree family contains active and passive indicative, wh-, relative clause, sentential adjunct and gerund trees. This set collapses into an active, two passive and one or two gerund categories. On the other hand a wh- word like *who*, which selects only a +WH NP and a relative pronoun tree in XTAG, maps to eight CCG categories. Given the space constraints here, we will only discuss the two primary differences we found in translating the XTAG grammar to CCG. The first difference involves the handling of extraction, and the second involves the treatment of VPs and non-constituents.

### 4.1 Extraction: Wh- Questions and Relative Clauses

The biggest change in converting the XTAG grammar to CCG is in the treatment of extraction. In XTAG, each type of wh- extraction and relativization (subject, object, indirect object) has its own tree in every appropriate tree family (where tree families contain all of the related clausal trees for a given subcategorization frame). Thus, the workload is borne by the verbs which anchor the trees. As noted above, in translation to CCG each tree family collapses into an active and a passive category, plus one or two gerund categories[5]. This consolidation is licensed by the shifting of responsibility for extraction away from the CCG category for the verb, and to the wh- words and relative pronouns. Rather than simply being +WH elements as they are in XTAG, wh- words in CCG are complex categories, which "absorb" the extracted argument of the verbal category they compose with. Relative pronouns operate similarly, but yield an NP modifier rather than an S. As a result, the CCG syntactic lexicon contains multiple categories for each wh- word, totaling 35, which replace 211 extraction trees.

---
[4] It is interesting to note that 72% of the XTAG grammar can be covered with about a third of the features. This suggests that many of the features handle specialized areas of the grammar.

### 4.2 Constituents Available in the CCG

The second major difference between the XTAG grammar and the CCG built from it is that the XTAG grammar as it stands does not have "non-constituents", or VP as a root category of any initial tree.[6] This means that the XTAG system can only handle coordination of complete constituents, excluding VPs. CCG does not face this problem. Coordination in CCG is executed via a schema over categories which allows any two like categories to combine; thus, we obtain wider coverage than the XTAG grammar without additional categories (in fact, we eliminate the few specific coordination trees XTAG uses). The CCG can handle VP coordination, such as (1), using only the coordination schema and can handle gapping, as in (2), by using the coordination schema and subject type-raising.

(1) Paddington makes marmalade sandwiches and eats them every day.

(2) Paddington loves and Betsy hates marmalade sandwiches.

In addition, the XTAG grammar treats all verbal complements as sentences, and allows empty subjects (PRO) for infinitival and gerundive clauses. It is more natural to handle these as VPs in CCG, with the category S\NP. As a result, the analysis of sentential subjects (Ssubjs) and sentential complements (Scomps) is somewhat different. In XTAG, both of these constructions use full sentences, with features controlling whether and which complementizers adjoin. There is no explicit notion of S-bar or CP. In the CCG analysis, each verb that takes both infinitival and indicative complements (or subjects) will have one additional category to handle infinitives, e.g. $(S\backslash NP)/(S_{inf}\backslash NP)$. This distinction is made via features in the XTAG analysis and does not require a separate tree. Furthermore, the CCG needs an additional feature to distinguish clauses with extraction or complementizers (we are using +BAR). In the XTAG grammar, this information is carried by the COMP feature, which was not used for the CCG. Thus, indicative Scomps with complementizers will have the feature +BAR in CCG; verbs which take *that* complements will also allow −BAR Scomps, since the complementizer is optional. Indicative Ssubjs will all be +BAR, as the complementizer is obligatory.

## 5 Type Raising

The question of where to do type-raising is a perennial problem for any CCG parser. If it is done by the parser, then type-raising is driven by the needs of the particular sentence being parsed and the lexicon can remain more compact. However, this merely shifts the burden to the parser, which must generate the necessary type-raised categories, and it may change the complexity of parsing if unbounded type-raising is allowed (see [3] on the issue of generalized type-raising). This option also exempts the raised categories from any pre-parse filtering techniques. A second possibility is to have all of the required type-raised categories in the lexicon. This, however, could result in a lexicon which is large, and cumbersome to work with. We have developed an alternate solution – "hidden" type-raising. The lexicon used for grammar development will contain only non-type-raised categories, but when the lexicon is converted to a database for use by the parser, the necessary raised categories will be automatically generated and added to the database. This will enable us to keep the working lexicon small, and yet will allow the parser to take advantage of the filters which weed out certain categories.

We have not yet implemented the type-raising component of the system, but we intend to fol-low the principle of adding only what we need to obtain the same level of coverage as provided by the source LTAG. Crucially, we need type-raised categories which allow non-subject extraction.

---

[5]The category for the gerundive form of a given verb is just an NP looking for whatever arguments the normal verb expects.

[6]Verbs project trees containing all of their arguments, including subjects. Since the XTAG grammar requires that all trees be "complete", i.e. have all of their argument slots filled, the grammar cannot do VP coordination without the addition of schemas over trees.

# 6 The Parser

As in any lexicalized grammar, the CCG parser in this system can be partitioned into two stages. The first stage selects categories associated with the words of the input from the syntactic database. The second stage combines the selected categories in the process of a derivation.

## 6.1 Category Selection

A number of mechanisms can be used in the first stage to minimize the number of categories sent on to the second stage of the parser. One technique is to use bottom-up information about the categories such as the number of arguments the category requires relative to the number of arguments that are present in the input, the position of the functor in the sentence and any other lexical constraints present in the category. A second technique that we intend to adapt from the XTAG system is Supertagging [5]. The initial statistics for this component will be collected from the XTAG-parsed corpus using the translations from trees to categories. Since the mapping is not one-to-one, the statistics will not be precise, but will nonetheless serve to speed up the CCG parser and enable us to collect more accurate statistics on actual CCG derivations. The supertagger selects the $n$ most likely categories for each word. These are then passed to the parser; any seldom used categories, potentially including type-raised ones, will be dispreferred but still available.

## 6.2 Category Combination

The second stage of the parser is a unification-based CKY-style parser [8]. The unification mechanism is based on an algorithm developed by Tomabechi [7]. This unification algorithm, called "quasi-destructive" unification, unifies two feature structures by making alterations to each structure which are marked as temporary. The result of unification is then obtained by copying either of the original structures, and making permanent the temporary alterations in the copy, while invalidating any alterations to the original structure.

# 7 Summary

In this paper we have presented a progress report on a wide-coverage CCG system being built from the XTAG system. Although work remains to be done in integrating the grammar and the parser, we are encouraged by the rapid progress made thus far. It is to be noted that in a matter of four months we have developed a CCG grammar to cover 70% of the XTAG grammar which has taken nearly six years to evolve [1, 2]. It is due to the close relationship of the two formalisms that we have been able to proceed so quickly. We hope that the working system will serve to further elucidate the relationship between CCG and LTAG. On the practical side the system is intended to serve as a CCG grammar development environment.

**Acknowledgments**: We would like to thank Julie Bourne, Hoa Dang, and David Fine for help in translating the trees. Al Kim wrote the parser and has also been helpful throughout the project. Mark Steedman has offered support and encouragement.